\documentclass{article}

\usepackage{microtype}
\usepackage{graphicx}
\usepackage{subcaption}
\usepackage{booktabs} 

\usepackage{hyperref}
\usepackage[table]{xcolor}

\usepackage[accepted]{icml2026}
\makeatletter
\renewcommand{\ICML@appearing}{%
  \textit{Accepted to the Structured Probabilistic Inference \& Generative Modeling workshop at ICML 2026.}%
}
\makeatother

\usepackage{amsmath}
\usepackage{amssymb}
\usepackage{mathtools}
\usepackage{amsthm}
\usepackage{enumitem}

\usepackage{pifont}
\newcommand{\cmark}{\ding{51}} 
\newcommand{\xmark}{\ding{55}} 
\usepackage[capitalize,noabbrev]{cleveref}

\theoremstyle{plain}

\theoremstyle{definition}

\theoremstyle{remark}

\usepackage[textsize=tiny]{todonotes}
\usepackage{multirow}
\usepackage{stfloats}
\icmltitlerunning{\texorpdfstring{PRISM-SLAM: Probabilistic Ray-Grounded Inference \\ for Scale-aware Metric SLAM}{PRISM-SLAM: Probabilistic Ray-Grounded Inference for Scale-aware Metric SLAM}}

\begin{document}

\twocolumn[
  \icmltitle{\texorpdfstring{PRISM-SLAM: Probabilistic Ray-Grounded Inference \\ for Scale-aware Metric SLAM}{PRISM-SLAM: Probabilistic Ray-Grounded Inference for Scale-aware Metric SLAM}}



  \begin{icmlauthorlist}
    \icmlauthor{Eunsoo Im}{km}
    \icmlauthor{Gyeonggwan Lee}{km}
    \icmlauthor{Seunghwan Hong}{km}
    \icmlauthor{Junghun Suh}{km}
  \end{icmlauthorlist}

  \icmlaffiliation{km}{KakaoMobility, South Korea}

  \icmlcorrespondingauthor{Eunsoo Im}{ies041196@gmail.com}


  \vskip 0.3in
]



\printAffiliationsAndNotice{}  

\icmltitlerunning{PRISM-SLAM}

\begin{abstract}
  Monocular SLAM historically suffers from scale ambiguity and tracking failure in dynamic environments.
  While recent vision foundation models (VFMs) provide remarkable zero-shot depth priors,
  naively integrating these deterministic predictions ignores
  predictive uncertainty and frame-to-frame scale inconsistencies.
  We propose PRISM-SLAM,
  a real-time framework that rigorously integrates VFM priors into a structured Bayesian factor graph
  to achieve scale-aware, metric-consistent localization and mapping.
  Specifically, we introduce a Pl\"ucker Ray-Distance Factor
  to anchor monocular observations in absolute space
  within a globally consistent metric coordinate system,
  constraining the scale-ambiguous monocular problem by integrating learned metric depth priors, thereby making the metric scale Fisher-identifiable.
  To handle environmental dynamics,
  we derive an epistemic uncertainty proxy from temporal depth consistency
  and formulate a Dynamic Scene Uncertainty Gating (DSUG) mechanism.
  This soft-gating approach probabilistically down-weights dynamic distractors
  without incurring the heavy computational overhead
  associated with traditional semantic segmentation masks.
  By employing a multi-process architecture
  that asynchronously processes VFM inference and geometric tracking,
  PRISM-SLAM provides verified metric output at 30 FPS using solely RGB input,
  bridging the gap between foundation models and real-world robotic applications.
  Evaluated on the TUM RGB-D and 7-Scenes benchmarks,
  PRISM-SLAM achieves a metric $SE(3)$ Absolute Trajectory Error (ATE)
  nearly identical to its oracle-aligned $Sim(3)$ error.
  This demonstrates that our system can produce deployment-ready metric trajectories
  by delivering robust metric SLAM solutions without any post-hoc scale correction.

  Project page: \url{https://ies0411.github.io/prism_slam_pj/}
\end{abstract}

\section{Introduction}

Monocular SLAM is a foundational technology for autonomous driving and robotics
due to its minimal hardware constraints.
However, traditional geometry-based systems \cite{campos2021orb} are bound
by a fundamental mathematical limitation:
the unobservability of absolute scale
due to the inherent ambiguity of pinhole projective geometry \cite{hartley2003multiple}.
This critical flaw leads to severe scale drift,
a problem further exacerbated in dynamic environments
where strict scene rigidity assumptions are routinely violated \cite{bescos2018dynaslam, zheng2025wildgs}.

Recently,
Vision Foundation Models (VFMs) have demonstrated remarkable potential
in reconstructing 3D structures from single images.
While these models offer a path toward metric-consistent SLAM,
existing learning-based systems \cite{teed2021droid, matsuki2024gaussian, lipson2024deep} often treat neural predictions as deterministic ground truth.
This naive integration ignores predictive uncertainty and frame-to-frame scale inconsistency, leading to optimization instabilities and degraded map quality.

Furthermore, many recent methods \cite{maggio2025vggt, zheng2025wildgs} exhibit distinct limitations that hinder real-world robotic deployment.
As surveyed in Table~\ref{tab:metric_survey}, even among recent monocular systems incorporating metric depth priors,
strict $SE(3)$ evaluation without post-hoc oracle correction remains largely unverified \cite{teed2021droid, deng2025gigaslam}.
Some suffer from severe computational latency that prevents real-time operation,
while others rely on post-hoc $Sim(3)$ alignment,
recovering an oracle scale factor from ground truth after the fact \cite{maggio2025vggt, zheng2025wildgs}.
Crucially, providing a verified,
metric-scale trajectory at runtime without oracle supervision remains an open challenge \cite{matsuki2024gaussian, murai2025mast3rslam}.

To bridge this gap, we propose PRISM-SLAM, a real-time framework
that treats foundation model outputs as probabilistic priors within a structured inference engine.
Rather than injecting deterministic depth directly,
our approach probabilistically fuses high-frequency geometric tracking
with uncertainty-aware VFM depth and rays.
By formulating metric predictions as orthogonal ray-distance constraints
and gating dynamic distractors using an epistemic uncertainty proxy,
we achieve robust metric consistency
without requiring explicit semantic segmentation or post-hoc alignment.
Our core contributions are:

\begin{itemize}
  \item Pl\"ucker Ray-Distance Factor:
        We introduce a 3D ray-distance formulation
        that anchors monocular observations in a globally consistent metric coordinate system.
        By furnishing explicit scale gradients,
        this factor effectively eliminates the rank-deficient null-space of standard 2D reprojection,
        thereby resolving fundamental scale ambiguity and drift.

  \item Dynamic Scene Uncertainty Gating (DSUG):
        We propose DSUG, a soft-gating mechanism
        that probabilistically filters dynamic distractors and unreliable depth regions.
        This ensures robust metric optimization in complex environments
        without the need for explicit semantic segmentation.

  \item Log-Domain Scale Adaptive Filter:
        We develop an asynchronous scale estimator
        to bridge the temporal gap between the real-time tracking frontend and the VFM-based backend.
        This ensures continuous, strictly positive metric scale feedback
        while maintaining the real-time efficiency of the multi-process architecture.

  \item ViT-Driven Metric Loop Fusion:
        We repurpose the foundation model's ViT tokens
        for zero-cost place recognition and global metric correction.
        This approach ensures topological consistency across large-scale trajectories
        and resolves accumulated drift without requiring complex map-merging.
\end{itemize}

\begin{table}[t]
  \centering
  \caption{
    \textbf{Metric scale capabilities of monocular SLAM systems.}
    While recent methods incorporate metric depth priors,
    they still rely on post-hoc $Sim(3)$ trajectory alignment.
    In SLAM evaluation,
    $Sim(3)$ alignment allows the trajectory to be rescaled using a ground-truth oracle,
    whereas $SE(3)$ alignment strictly evaluates rigid body transformations
    without any scale correction.
    PRISM-SLAM achieves true metric tracking evaluated under strict $SE(3)$ alignment by incorporating learned metric priors to constrain the monocular geometry.
    (n.s. stands for not specified).
  }
  \label{tab:metric_survey}
  \resizebox{\columnwidth}{!}{%
    \begin{tabular}{llcc}
      \toprule
      Method                                 & Depth Prior                            & Alignment      & FPS         \\
      \midrule
      ORB-SLAM3 \cite{campos2021orb}         & --                                     & Sim(3)         & 30          \\
      DROID-SLAM \cite{teed2021droid}        & --                                     & Sim(3)         & 5           \\
      DPV-SLAM++ \cite{lipson2024deep}       & --                                     & Sim(3)         & 50          \\
      GO-SLAM \cite{zhang2023goslam}         & --                                     & Sim(3)         & 3           \\
      MonoGS \cite{matsuki2024gaussian}      & --                                     & Sim(3)         & 3           \\
      Splat-SLAM \cite{sandstrom2025splat}   & Omnidata \cite{eftekhar2021omnidata}   & Sim(3)         & 1.2         \\
      MASt3R-SLAM \cite{murai2025mast3rslam} & MASt3R \cite{leroy2024grounding}       & Sim(3)         & 15          \\
      VGGT-SLAM \cite{maggio2025vggt}       & VGGT  \cite{wang2025vggt}              & Sim(3)         & 20          \\
      EC3R-SLAM \cite{hu2025ec3rslam}        & VGGT \cite{wang2025vggt}               & Sim(3)         & 36          \\
      WildGS-SLAM \cite{zheng2025wildgs}     & Metric3D v2 \cite{hu2024metric3d}      & Sim(3)         & 0.5         \\
      GigaSLAM \cite{deng2025gigaslam}       & UniDepth \cite{piccinelli2024unidepth} & n.s.           & --          \\
      \midrule
      \textbf{PRISM (Ours)}                  & \textbf{DA3} \cite{yang2025depth3}     & \textbf{SE(3)} & \textbf{30} \\
      \bottomrule
    \end{tabular}%
  }
\end{table}

\section{Related Works}
\subsection{Monocular SLAM in Dynamic Environments}

Traditional monocular SLAM systems, most notably ORB-SLAM3 \cite{campos2021orb},
have established robust standards through multi-map management and tight bundle adjustment (BA).
However, these systems fundamentally rely on scene rigidity assumptions,
making them vulnerable to tracking errors in environments where objects move independently.
To address dynamic distractors,
DynaSLAM \cite{bescos2018dynaslam} incorporated semantic segmentation masks,
while DROID-SLAM \cite{teed2021droid} leveraged dense recurrent optical flow
for robust correspondence estimation.
While effective, these methods incur severe computational costs and
often struggle to generalize across diverse, unpredictable motion patterns in the wild.

Beyond traditional sparse tracking,
recent advancements in dense SLAM have shifted the focus toward high-fidelity volumetric maps
using novel view synthesis representations
like Neural Radiance Fields (NeRFs) and 3D Gaussian Splatting (3DGS) \cite{kerbl20233d}.
While systems like MonoGS \cite{matsuki2024gaussian} were early pioneers in utilizing 3DGS,
they typically assume static environments, leading to significant artifacts in the presence of motion.
To mitigate this, WildGS-SLAM \cite{zheng2025wildgs} introduced an uncertainty-aware pipeline
using DINOv2 features and a shallow MLP to predict dynamic masks.
However, its reliance on pixel-wise hard masking and incremental MLP training creates
optimization gradient discontinuities and severe bottlenecks,
limiting operation to a non-real-time $\sim$0.5~FPS.
PRISM-SLAM elegantly bypasses these bottlenecks
by introducing Dynamic Scene Uncertainty Gating (DSUG) instead of explicit masks,
achieving real-time 30~FPS tracking.

\subsection{Scale Ambiguity and Vision Foundation Models}

Beyond dynamic distractors,
a critical challenge in monocular SLAM is the inherent scale ambiguity of pinhole projective geometry,
which fundamentally limits 3D reconstruction to an estimate
defined only up to a 7-DOF similarity transform
---or a 15-DOF projective transform in uncalibrated scenarios---
thereby lacking absolute metric scale.
VGGT-SLAM \cite{maggio2025vggt} addressed the projective distortion
by optimizing submap alignment on the $SL(4)$ manifold,
the Special Linear group representing 3D homographies.
However, while this manifold optimization mitigates projective distortions,
it incurs severe GPU memory overheads and
still fundamentally fails to recover absolute metric scale using pure geometry.
To break this inherent geometric bottleneck and directly inject real-world scale into the system,
the field has recently turned to deep structural priors.
Specifically, Vision Foundation Models (VFMs)
like Metric3D v2 \cite{hu2024metric3d} and Depth Anything \cite{yang2024depth}
have revolutionized zero-shot metric depth estimation.

Despite the availability of these metric priors,
existing learning-based SLAM systems struggle to provide verified metric output.
For instance,
WildGS-SLAM \cite{zheng2025wildgs} utilizes Metric3D v2 but
evaluates tracking only with post-hoc $Sim(3)$ alignment,
failing to verify metric trajectory accuracy at runtime.
PRISM-SLAM bridges this gap
by being the first to mathematically integrate
cross-view consistent metric rays from the DA3 foundation model,
alongside our epistemic uncertainty proxy,
directly into the Bayesian factor graph.
This formulation allows our system to resolve
both scale ambiguity and projective distortion in real-time,
delivering deployment-ready metric trajectories
without requiring post-hoc oracle alignment.

\section{Methodology}

\begin{figure*}[!ht]
  \centering
  \includegraphics[width=1.0\textwidth]{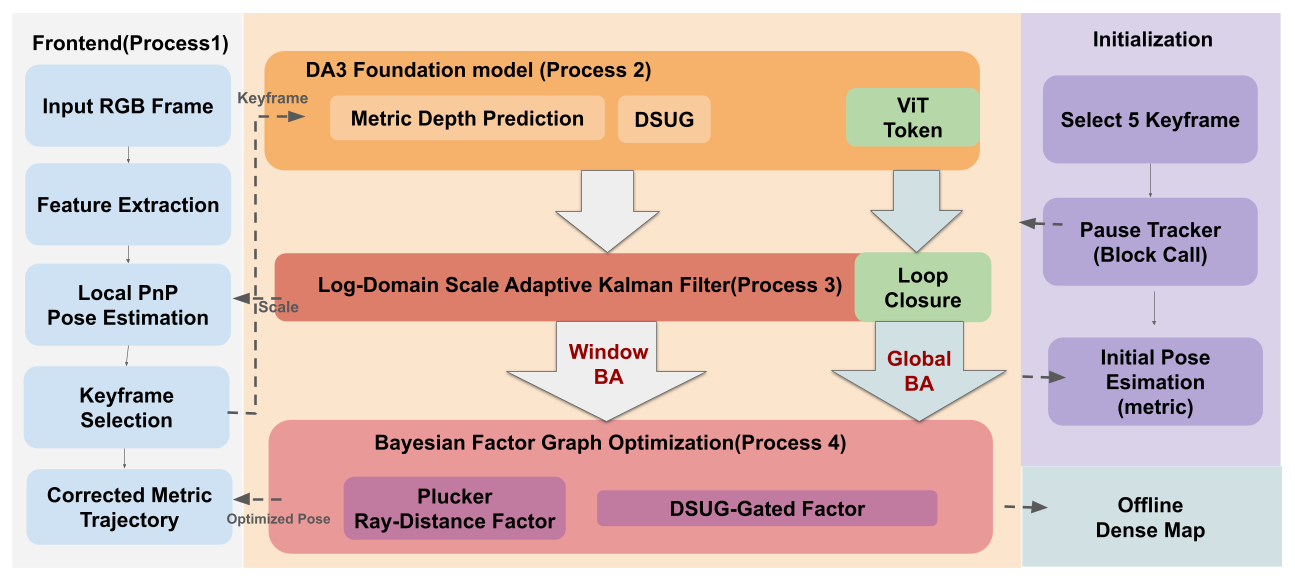}
  \caption{
    \textbf{PRISM-SLAM system architecture.}
    Our decoupled pipeline operates across four concurrent processes.
    \textbf{(1) Tracking:}
    A CPU-based frontend ($\sim$30~Hz) estimates initial poses and sparse points.
    \textbf{(2) VFM Extraction:}
    An asynchronous GPU worker extracts dense metric depth and uncertainty priors via DA3.
    \textbf{(3) Scale Recovery (KF):}
    A log-domain Kalman filter and WLS estimator dynamically fuse VFM priors with sparse points
    to resolve the monocular scale ambiguity.
    \textbf{(4) Metric Graph Optimization:}
    The backend refines the trajectory and map,
    employing the DSUG gate to filter dynamic artifacts and enforce metric consistency.
  }
  \vspace{-2mm}
  \label{fig:pipeline}
\end{figure*}

The overall pipeline of PRISM-SLAM is illustrated in Figure \ref{fig:pipeline}.
To maximize computational efficiency while maintaining high-fidelity metric accuracy,
our system operates on an asynchronous multi-process architecture.
The high-frequency frontend performs real-time tracking,
while the backend worker concurrently handles metric depth estimation
via the DA3 foundation model and graph optimization.
This decoupled design allows PRISM-SLAM to provide stable,
metric-scale updates without bottlenecking the real-time tracking loop.

Within this architecture, we first analyze the inherent scale ambiguity of monocular SLAM (\ref{sec3.1}).
To resolve this, we introduce Dynamic Scene Uncertainty Gating (DSUG) (\ref{subsec_DSUG}),
a probabilistic soft-gating mechanism that filters out dynamic distractors from the optimization objective.
To bridge the temporal gap between the asynchronous processes,
we deploy a Log-Domain Kalman Filter (\ref{subsec_AF}) that provides continuous,
strictly positive scale feedback to the tracker.

The core of our metric consistency lies in Metric Graph Optimization (\ref{subsec_MGO}),
where we introduce a Pl\"ucker Ray-Distance Factor to render the absolute scale Fisher-identifiable.
To ensure a near-perfect metric baseline from the start,
we utilize a Metric-Aware Initialization strategy (\ref{subsec_MAI}).
Finally, the backend ensures global topological consistency through ViT-Driven Loop Closure (\ref{subsec_ViT_LC})
before generating a high-fidelity Offline Global Metric Reconstruction (\ref{subsec_OGR}).

\subsection{The Scale Ambiguity Problem}
\label{sec3.1}
Standard monocular SLAM \cite{campos2021orb} extracts sparse features \cite{rublee2011orb},
estimates poses via PnP-RANSAC \cite{lepetit2009epnp, fischler1981random},
and refines landmarks through bundle adjustment \cite{triggs1999bundle}.
Its core limitation is the reliance on the 2D reprojection error between an observed pixel $p_{ij}$
and the projection $\pi$ of a 3D landmark $X_j$ transformed by the camera pose $T_i \in SE(3)$:
\begin{equation}
  E = \sum_{i,j} \rho \left( || p_{ij} - \pi(K (R_i X_j + t_i)) ||^2_{\Sigma_{ij}} \right)
\end{equation}
where $K$ is the intrinsic matrix,
$\Sigma_{ij}$ denotes the observation information matrix (the inverse of the measurement covariance)
that weights the residual components,
and $\rho$ is a robust loss function.
The term $||\cdot||^2_{\Sigma_{ij}}$ represents the squared Mahalanobis distance,
which accounts for the anisotropic noise distribution in the image plane.
The critical flaw in this formulation is the inherent scale ambiguity \cite{hartley2003multiple}.
If we multiply the camera translation and the landmark coordinates by an arbitrary positive scalar $s > 0$,
the projected 2D coordinates remain mathematically identical:
\begin{equation}
  \begin{split}
    \pi(K (R_i (s \cdot X_j) + s \cdot t_i)) & = \pi(s \cdot K (R_i X_j + t_i))              \\
                                             & = \pi(K (R_i X_j + t_i)), \quad \forall s > 0
  \end{split}
\end{equation}

Since the common scale factor is eliminated during projective division (z-normalization),
the Hessian matrix associated with the reprojection objective exhibits
a rank-deficient null-space along the scale dimension.
Consequently, the optimization engine receives no gradient information
to constrain or correct scale drift, resulting in an unobservable metric scale
that can fluctuate unpredictably over extended trajectories.


\subsection{Dynamic Scene Uncertainty Gating (DSUG)}
\label{subsec_DSUG}

\begin{figure*}[t]
  \centering
  \includegraphics[width=0.90\linewidth]{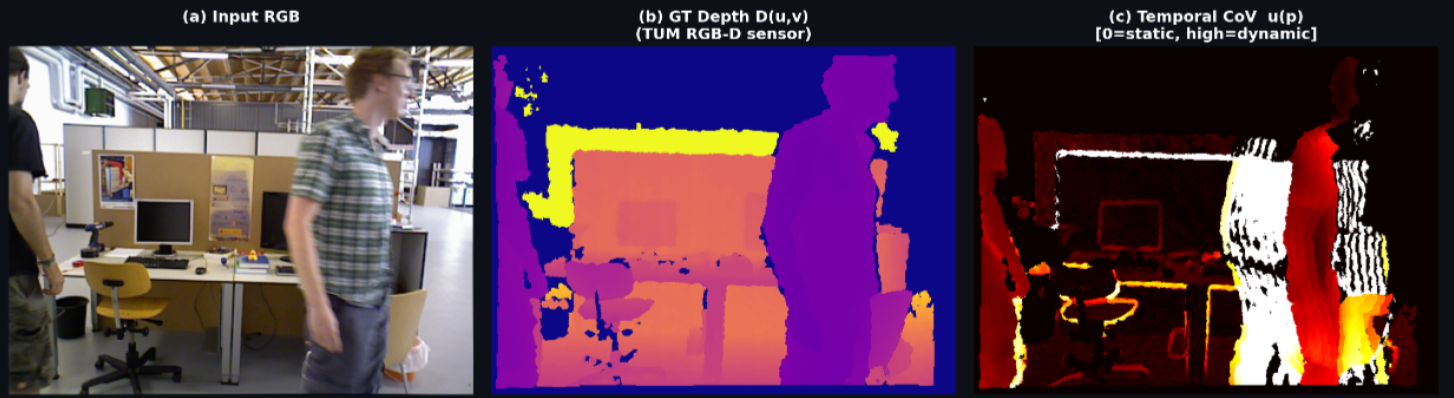}

  \caption{\textbf{Temporal Uncertainty Modeling in Dynamic Scenes.}
    \textbf{(a)} Input RGB frame from the TUM RGB-D \texttt{fr3/walking\_static} sequence.
    \textbf{(b)} Ground-truth depth map.
    \textbf{(c)} Pose-compensated depth residual of DA3 estimates, utilized as our DSUG epistemic uncertainty proxy $u(p)$.
    Bright regions indicate high temporal depth variation,
    precisely capturing the geometrically unstable boundaries of moving subjects.
    By mapping this variance directly to the optimization information matrix,
    PRISM-SLAM naturally down-weights the most disruptive dynamic regions without relying on hard semantic masks.}
  \label{fig:f1_dsug}
\end{figure*}

In dynamic environments, moving objects severely corrupt tracking geometry.
Existing methods attempt to handle this using binary semantic masks \cite{zheng2025wildgs},
but hard thresholding abruptly severs optimization edges,
causing gradient discontinuities
that disrupt non-linear solvers like Levenberg-Marquardt \cite{levenberg1944method}.
To overcome the optimization failures caused by these hard boundaries,
PRISM-SLAM introduces Dynamic Scene Uncertainty Gating (DSUG).
Because standard foundation models like DA3 \cite{yang2025depth3}
output deterministic depths without a native variance head,
we construct a hybrid epistemic uncertainty proxy $u(p)$
that fuses two complementary sources:
\begin{equation}  \label{eq:dsug_proxy}
  u(p) = \alpha \cdot u_{spatial}(p) + (1-\alpha) \cdot u_{temporal}(p)
\end{equation}
where $\alpha \in [0,1]$ is a balancing parameter,
and $p \in \mathbb{R}^2$ denotes a pixel location in the image.
Here, $u_{spatial}(p)$ is derived by inverting DA3's native spatial confidence map
—which naturally drops around the blurred or ambiguous boundaries of moving objects—
to capture predictive ambiguity.
Complementarily,
$u_{temporal}(p)$ denotes the depth discrepancy between consecutive keyframes
after compensating for camera ego-motion,
which explicitly captures geometric instability over time,
as visualized in Figure \ref{fig:f1_dsug}.

Instead of applying a binary cutoff,
we map this combined uncertainty into a continuous precision weight for the Bayesian Information Matrix ($\Omega$).
To prevent numerical instability (i.e., avoiding $\Omega \to \infty$ when $u(p)$ approaches zero),
we formulate a probabilistic soft-gating mechanism
using a bounded sigmoid function:
\begin{equation} \label{eq:dsug}
  w(p) = \sigma\left(\frac{\tau - u(p)}{T}\right), \quad \Omega(p) = \frac{w(p)}{\sigma_0^2}
\end{equation}
where $w(p) \in [0, 1]$ represents the normalized gating weight for a pixel $p$,
$\tau$ denotes the uncertainty threshold
that defines the decision boundary between static and dynamic regions,
and $T$ is the temperature parameter controlling the smoothness of the gate's transition.
The term $\sigma_0^2$ serves as a fixed calibration constant
representing the nominal measurement noise of the sensor.
The output of our probabilistic soft-gating mechanism,
$\Omega(p)$, represents a precision weight in the information matrix,
which gracefully down-weights dynamic distractors during optimization.


\subsection{Log-Domain Scale Adaptive Filter}
\label{subsec_AF}

Estimating a consistent global scale $s$ from the foundation model
presents two primary mathematical challenges.
First, varying network confidence induces heteroscedastic noise in the scale observation.
Second, scale is a strictly positive, multiplicative quantity ($s > 0$),
Standard linear estimators operating in Euclidean space assume additive noise,
which can theoretically yield physically impossible negative scale values,
thus violating this strict constraint.

To resolve this, we map the scale variable to the log-domain ($\log s$),
elegantly transforming multiplicative scale drift into additive noise.
Within this log-space, we compute the single-frame scale observation
using a Weighted Least Squares (WLS) formulation.
Crucially, this WLS directly ingests the DSUG information matrix $\Omega$
derived in Section \ref{subsec_DSUG} as its precision weights.
By naturally down-weighting dynamic distractors and ambiguous pixels,
the WLS gracefully filters out the heteroscedastic network noise.
Finally, tracking these robust log-scale observations with a 1D Kalman filter,
using the WLS variance as adaptive measurement noise,
guarantees a smooth and strictly positive metric consistency.
The refined scale is then continuously fed back to the frontend
to correct real-time pose estimation and
forwarded to the backend to constrain the Window BA.

\subsection{Metric Graph Optimization}
\label{subsec_MGO}

To retroactively correct the historical trajectory and map points,
the backend performs covisibility-based Bundle Adjustment (BA).
This optimization operates locally over a spatial window during standard tracking,
and scales globally across the entire graph upon loop closure detection.
Regardless of the optimization scope,
rigorously locking the graph to the absolute metric scale
requires breaking the scale null-space of standard 2D reprojection.

To achieve this,
we introduce a 3D Pl\"ucker Ray-Distance formulation.
By lifting 2D pixels using the VFM's metric depth predictions,
we project the network's spatial targets into the 3D world as anchored, infinite rays.
For a metric 3D direction vector $d_i$,
the ray emanating from the camera center $t_i$ is parameterized
by 6D Pl\"ucker coordinates $L = (d_i, m_i)$ \cite{bartoli2005structure},
where the moment vector $m_i = t_i \times d_i$ is computed
using the cross-product operator.
The orthogonal distance residual for a 3D landmark $X_k$ is formulated as:

\begin{equation} \label{eq:plucker_residual}
  e_{\text{ray}}(T_i, X_k) = \frac{\|d_i \times X_k + m_i\|}{\|d_i\|}
\end{equation}
where $T_i$ represents the camera pose,
$d_i$ is the unit direction vector of the ray,
$m_i$ is the Pl\"ucker moment, and $X_k$ is the 3D landmark position in the world frame.
The numerator $\|d_i \times X_k + m_i\|$ computes the magnitude of the moment of $X_k$
with respect to the ray,
effectively measuring the perpendicular distance from the point to the infinite line.
Because the VFM's metric prediction rigidly anchors the ray
in a globally consistent coordinate system,
applying an arbitrary global rescaling $s \cdot X_k$ forces
the landmark to physically deviate from the ray,
strictly increasing the residual.
While a single point-to-ray distance residual along a ray does not uniquely fix the absolute depth on its own,
fusing these orthogonal constraints across multiple views---in conjunction with the continuous scale feedback from our log-domain Kalman filter---tightly anchors the global metric scale.
This joint optimization renders the scale locally Fisher-identifiable \cite{huang2009observability} and successfully resolves accumulated scale drift.

Crucially, to prevent dynamic objects and erroneous depth predictions
from corrupting the map geometry,
we employ a DSUG-Gated weighting mechanism.
The Pl\"ucker ray residuals are dynamically weighted
by the DSUG confidence map derived in Section \ref{subsec_DSUG}.
By naturally down-weighting ambiguous or moving distractors,
the graph optimization ensures that the metric anchoring is driven exclusively
by highly reliable, static scene structures.

\subsection{Metric-Aware Initialization}
\label{subsec_MAI}

Standard monocular SLAM initializes at an arbitrary scale,
leading to a significant temporal lag before reaching metric consistency.
PRISM-SLAM resolves this through a synchronous metric initialization strategy.
During the first few keyframes (e.g., $N_{\text{init}} = 5$),
the tracking frontend briefly suspends execution to await the DA3 depth.

This short synchronous period allows the system
to compute a robust initial scale $\hat{s}_0$ via log-domain WLS
and immediately apply strong metric constraints to the first map points.
The resulting estimate is used to warm-start the Kalman filter state,
providing a near-perfect metric baseline from the very first meter of trajectory.
Once the initial scale is locked,
the system seamlessly transitions to its standard asynchronous multi-process mode.
However, if the initial camera motion lacks sufficient translation
for robust triangulation (e.g., pure rotation),
this creates a degenerate geometric configuration.
To prevent initialization failure in such cases,
the system dynamically extends the synchronization window
until a stable geometric baseline is established.

\subsection{ViT-Driven Loop Closure and Metric Global BA}
\label{subsec_ViT_LC}
PRISM-SLAM repurposes the \texttt{[CLS]} token of the DA3 ViT backbone \cite{dosovitskiy2020image}
as a global scene descriptor for loop detection,
completely replacing traditional DBoW2 \cite{galvez2012bags} vocabularies
at zero additional computational cost.
Candidate loops are retrieved via cosine similarity
and geometrically verified by estimating the Essential matrix
within a RANSAC framework \cite{hartley2003multiple, fischler1981random}.

Upon confirmation,
a Metric Global BA jointly optimizes standard reprojection edges
alongside our proposed Pl\"ucker ray-distance and metric depth factors.
Crucially, the depth constraints act as absolute metric anchors,
guaranteeing strict scale consistency through the loop correction.
Simultaneously, the DSUG formulation probabilistically excludes dynamic distractors from the graph.
This comprehensive optimization ensures
that the final loop-corrected trajectory is metrically accurate,
topologically robust, and inherently free of dynamic artifacts.

\subsection{Offline Global Metric Reconstruction}
\label{subsec_OGR}

Since dense mapping is not the primary focus of the real-time tracking loop,
PRISM-SLAM adopts a decoupled, offline reconstruction strategy
to maximize final geometric consistency without bottlenecking the frontend.

Once the online SLAM trajectory is fully optimized via Global BA,
we collect the globally consistent metric keyframe poses.
Instead of naively accumulating noisy single-view depth predictions online,
we input batches of these optimized poses directly into DA3's multi-view inference module.
Guided by these accurate geometric anchors,
the network's cross-view attention mechanism resolves spatial inconsistencies,
generating highly coherent metric depth maps.

Concurrently, the previously computed DSUG masks are applied to filter out
dynamic distractors and transient occlusions directly
from these depth maps prior to 3D integration.
These refined, cross-view consistent depth maps are then directly
back-projected to construct a clean, global dense point cloud.
This decoupled architecture elegantly separates the real-time robustness
required for tracking from the heavy computational demands of high-fidelity dense mapping.

\section{Experiments}

The system was evaluated strictly using monocular RGB input on an NVIDIA RTX 4500 Ada GPU. We benchmarked on TUM RGB-D \cite{sturm12iros}, 7-Scenes \cite{shotton2013scene},
and BONN Dynamic \cite{palazzolo2019refusion}.
We report Absolute Trajectory Error (ATE) RMSE in centimeters using both Sim(3) alignment (standard) and, uniquely for PRISM-SLAM,
SE(3) alignment with the system's own metric scale. All results report the median of 3 independent runs unless noted.

\subsection{Metric Scale Recovery}

A defining capability of PRISM-SLAM is its ability to output metric trajectories in real-time, without any post-hoc scale correction.
We evaluate this by reporting SE(3) ATE,
rigid alignment without any scale degree of freedom, alongside the standard Sim(3) ATE.
On \texttt{fr1/xyz} (Table~\ref{tab:tum_fr1}), the SE(3) ATE closely matches the Sim(3) ATE (3.04~cm vs. 2.86~cm) with only a 3.3\% scale error,
demonstrating that the system's online metric scale recovery is so accurate that it is virtually identical to the oracle-aligned $Sim(3)$ solution,
which requires ground-truth knowledge.
Furthermore, this strict metric fidelity extends to the \texttt{fr3} dataset (Table~\ref{tab:tum_fr3}).
In static scenes, the SE(3) ATE remains tightly bound to the Sim(3) ATE (e.g., 1.8~cm vs. 1.6~cm on \texttt{sit}), proving robust scale consistency.
Even in highly dynamic environments (\texttt{sit-xyz} and \texttt{walk-xyz}) where scale recovery is notoriously vulnerable to moving distractors,
PRISM-SLAM successfully maintains a verified metric scale without relying on any offline scaling.


\begin{table}[t]
  \centering
  \caption{
    \textbf{ATE RMSE (cm) on TUM RGB-D \cite{sturm12iros} (fr1 Sequences).}
    All baselines are evaluated using standard Sim(3) alignment. For PRISM-SLAM, we additionally report the SE(3) ATE (Sim(3) \textit{/ SE(3)}),
    demonstrating strict metric tracking without oracle correction.
  }
  \label{tab:tum_fr1}
  \resizebox{\columnwidth}{!}{%
    \begin{tabular}{lcccc}
      \toprule
      \multirow{2}{*}{Method}                      & \multirow{2}{*}{FPS} & \multirow{2}{*}{Metric} & \multicolumn{2}{c}{fr1 Sequences (Sim(3) \textit{/ SE(3)})}               \\
      \cmidrule(lr){4-5}
                                                   &                      &                         & xyz                                                         & rpy         \\
      \midrule
      \multicolumn{5}{l}{\textit{RGB (Calibrated)}}                                                                                                                             \\
      ORB-SLAM3 \cite{campos2021orb}               & 30                   & \xmark                  & 0.9                                                         & --          \\
      DeepV2D \cite{teed2018deepv2d}               & 2                    & \xmark                  & 6.4                                                         & 10.5        \\
      DeepFactors \cite{czarnowski2020deepfactors} & 30                   & \xmark                  & 3.5                                                         & 4.3         \\
      DPV-SLAM \cite{lipson2024deep}               & 15                   & \xmark                  & 1.0                                                         & 3.0         \\
      DPV-SLAM++ \cite{lipson2024deep}             & 50                   & \xmark                  & 1.0                                                         & 3.2         \\
      GO-SLAM \cite{zhang2023goslam}               & 3                    & \xmark                  & 1.0                                                         & 1.9         \\
      DROID-SLAM \cite{teed2021droid}              & 5                    & \xmark                  & 1.2                                                         & 2.6         \\
      MASt3R-SLAM \cite{murai2025mast3rslam}       & 15                   & \xmark                  & 0.9                                                         & 2.7         \\
      \midrule
      \multicolumn{5}{l}{\textit{RGB (Uncalibrated)}}                                                                                                                           \\
      VGGT-SLAM \cite{maggio2025vggt}             & 20                   & \xmark                  & 1.4                                                         & 3.0         \\
      \textbf{PRISM (Ours)}                        & 30                   & \cmark                  & 2.86 / 3.04                                                 & 4.10 / 4.94 \\
      \bottomrule
    \end{tabular}%
  }
  \vspace{1mm}
\end{table}

\begin{table}[t]
  \centering
  \caption{
    \textbf{ATE RMSE (cm) on TUM RGB-D \cite{sturm12iros} (fr3 Sequences).}
    Static and dynamic sequences.
    All baselines are evaluated using standard Sim(3) alignment. For PRISM-SLAM,
    we additionally report the SE(3) ATE (Sim(3) \textit{/ SE(3)}),
    demonstrating strict metric tracking without oracle correction.
  }
  \label{tab:tum_fr3}
  \resizebox{\columnwidth}{!}{%
    \begin{tabular}{lccccc}
      \toprule
      \multirow{2}{*}{Method}            & \multirow{2}{*}{Metric} & \multicolumn{2}{c}{Static (Sim(3) \textit{/ SE(3)})} & \multicolumn{2}{c}{Dynamic (Sim(3) \textit{/ SE(3)})}                             \\
      \cmidrule(lr){3-4} \cmidrule(lr){5-6}
                                         &                         & sit                                                  & walk                                                  & sit-xyz     & walk-xyz    \\
      \midrule
      \multicolumn{6}{l}{\textit{RGB (Calibrated)}}                                                                                                                                                           \\
      ORB-SLAM3 \cite{campos2021orb}     & \xmark                  & 0.7                                                  & 0.9                                                   & 25.3        & 48.8        \\
      DROID-SLAM \cite{teed2021droid}    & \xmark                  & 0.4                                                  & 0.3                                                   & 17.6        & 21.4        \\
      MonoGS \cite{matsuki2024gaussian}  & \xmark                  & 1.1                                                  & 3.6                                                   & 16.2        & 18.9        \\
      WildGS-SLAM \cite{zheng2025wildgs} & \xmark                  & 0.5                                                  & 0.6                                                   & 4.1         & 5.2         \\
      \midrule
      \multicolumn{6}{l}{\textit{RGB (Uncalibrated)}}                                                                                                                                                         \\
      \textbf{PRISM (Ours)}              & \cmark                  & 1.6 / 1.8                                            & 1.9 / 2.7                                             & 12.7 / 20.0 & 23.2 / 26.8 \\
      \bottomrule
    \end{tabular}%
  }
  \vspace{1mm}
\end{table}

\subsection{Tracking Accuracy}

\paragraph{Static TUM Sequences.}
On static sequences (Table~\ref{tab:tum_fr1}),
PRISM-SLAM achieves 2.86~cm $Sim(3)$ ATE on \texttt{fr1/xyz},
with the $SE(3)$ ATE closely following at 3.04~cm---confirming metric-scale fidelity.
Our system runs at 30~FPS on a single GPU, significantly faster than offline 3DGS optimization methods like WildGS-SLAM \cite{zheng2025wildgs} ($\sim$0.5~FPS).
On \texttt{fr3} static scenes (Table~\ref{tab:tum_fr3}), PRISM achieves 1.6~cm and 1.9~cm $Sim(3)$ ATE on \texttt{sit} and \texttt{walk-static} respectively,
approaching dense methods like DROID-SLAM while uniquely providing strict metric output.

As noted, state-of-the-art baselines like ORB-SLAM3 or DROID-SLAM may yield lower $Sim(3)$ ATE through pure geometric optimization or intensive optical flow. However, their metric scale remains unobservable without oracle alignment.
We emphasize that the primary contribution of PRISM-SLAM is its metric output capability---providing deployment-ready $SE(3)$ trajectories at runtime---rather than claiming strictly superior absolute trajectory accuracy over $Sim(3)$-aligned baselines.

\paragraph{Dynamic TUM Sequences.}
Crucially, in highly dynamic environments (Table~\ref{tab:tum_fr3}, \texttt{sit-xyz} and \texttt{walk-xyz}),
PRISM-SLAM maintains robust tracking in challenging scenarios
where recent VFM-integrated SLAM systems exhibit degraded performance.
Notably, uncalibrated foundation-model-driven systems such as VGGT-SLAM \cite{maggio2025vggt} completely fail on these dynamic \texttt{fr3} sequences due to catastrophic tracking divergence.
This explicit failure case highlights the necessity of our DSUG formulation,
which elegantly marginalizes dynamic occlusions to preserve trajectory stability without relying on hard semantic masks.

\begin{table}[t]
  \centering
  \caption{
    \textbf{ATE RMSE (cm) on BONN Dynamic \cite{palazzolo2019refusion}.}
    All baselines utilize active RGB-D hardware for absolute scale, whereas PRISM-SLAM operates strictly on Monocular RGB.
    `-' indicates sequences not evaluated by the baselines.
  }
  \label{tab:bonn}
  \resizebox{\columnwidth}{!}{%
    \begin{tabular}{lccccc}
      \toprule
      \multirow{2}{*}{Method}               & \multirow{2}{*}{Metric} & \multicolumn{4}{c}{Dynamic Sequences (Sim(3) \textit{/ SE(3)})}                                            \\
      \cmidrule(lr){3-6}
                                            &                         & balloon                                                         & balloon2    & pers\_trk   & balloon\_trk \\
      \midrule
      \multicolumn{6}{l}{\textit{RGB-D (Hardware Scale)}}                                                                                                                          \\
      ORB-SLAM3 \cite{campos2021orb}        & \cmark                  & 5.8                                                             & 17.7        & 70.7        & -            \\
      DynaSLAM \cite{bescos2018dynaslam}    & \cmark                  & 3.0                                                             & 2.9         & 6.1         & -            \\
      ReFusion \cite{palazzolo2019refusion} & \cmark                  & 17.5                                                            & 25.4        & 28.9        & -            \\
      RoDyn-SLAM \cite{jiang2024rodyn}      & \cmark                  & 7.9                                                             & 11.5        & 14.5        & -            \\
      \midrule
      \multicolumn{6}{l}{\textit{RGB (Uncalibrated)}}                                                                                                                              \\
      \textbf{PRISM (Ours)}                 & \cmark                  & 9.8 / 18.1                                                      & 14.0 / 17.7 & 36.7 / 39.5 & 7.8 / 9.1    \\
      \bottomrule
    \end{tabular}%
  }
\end{table}

\subsection{Dynamic Tracking on BONN Dataset}

We further evaluate PRISM-SLAM on the Bonn Dynamic dataset, which presents highly challenging scenarios featuring large-scale moving objects that frequently occlude the static background.
As shown in Table~\ref{tab:bonn}, we report both $Sim(3)$ and $SE(3)$ ATE to provide a transparent assessment of our true metric tracking capabilities.

Notably, while the baseline methods benefit from active RGB-D hardware to obtain absolute scale, PRISM-SLAM recovers metric scale strictly from monocular RGB input.
On the \texttt{balloon2} and \texttt{pers\_trk} sequences, our $SE(3)$ errors (17.7~cm and 39.5~cm) closely align with their $Sim(3)$ counterparts (14.0~cm and 36.7~cm).
This tight alignment demonstrates that our deep structural priors and DSUG mechanism preserve a consistent metric scale even under severe dynamic interference.
Furthermore, while the $Sim(3)$ evaluations show that our purely monocular approach does not strictly surpass hardware-assisted RGB-D baselines like ReFusion \cite{palazzolo2019refusion} or DynaSLAM \cite{bescos2018dynaslam} in absolute accuracy, PRISM-SLAM still yields a highly competitive trajectory structure.
Bridging this performance gap without relying on active depth sensing underscores the robustness and practical viability of our real-time metric scale recovery.

\paragraph{7-Scenes Indoor Localisation.}
We additionally evaluate on the 7-Scenes benchmark \cite{shotton2013scene} to demonstrate our system's tracking robustness in texture-poor environments.
While extreme rotational motions in scenes like \texttt{heads} challenge pure monocular metric scale recovery (resulting in scale drift, see Appendix B),
the underlying trajectory geometry remains highly accurate. Utilizing a learned feature frontend (KeyNet),
our system yields an impressive mean Sim(3) ATE of 8.8~cm across all scenes.
This confirms that PRISM-SLAM maintains robust topological tracking even when ideal metric conditions are not met.


\begin{figure}[t]
  \centering
  \includegraphics[width=\columnwidth]{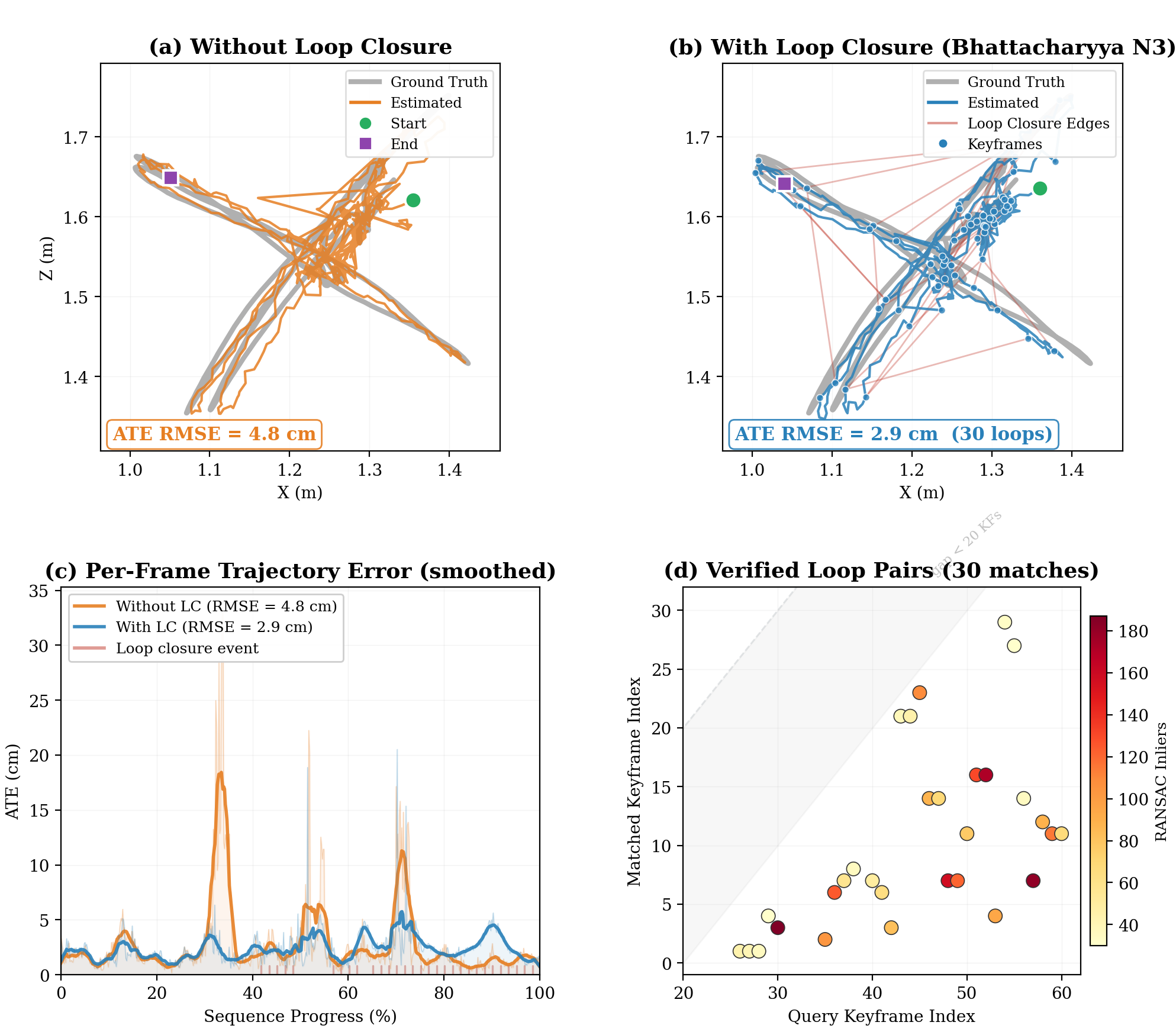}
  \caption{
    \textbf{Impact of ViT-driven Loop Closure on the TUM \texttt{fr1/xyz} sequence.}
    \textbf{(a) Without Loop Closure:} The purely visual odometry estimate (orange) progressively deviates from the ground truth (grey) due to accumulated scale and rotation drift,
    resulting in an ATE RMSE of 4.8 cm.
    \textbf{(b) With Loop Closure:} ViT-driven place recognition successfully detects 30 valid loops.
    Applying these geometric constraints (pink edges) globally optimizes the pose graph, tightly aligning the estimated trajectory (blue) with the ground truth and significantly reducing the ATE RMSE to 2.9 cm.
    \textbf{(c) Per-Frame Trajectory Error:} A temporal comparison of tracking errors. The severe error spikes in the uncorrected baseline (orange) are effectively neutralized by our system (blue).
    The red vertical markers at the bottom represent the exact timestamps of loop closure events, which explicitly coincide with immediate error drops.
    \textbf{(d) Verified Loop Pairs:} The temporal distribution of the 30 accepted matches (Query vs. Matched keyframe indices).
    The off-diagonal scattered points demonstrate the system's ability to recognize previously visited locations across large temporal gaps.
    The color map denotes the number of RANSAC inliers, reflecting the high geometric confidence of the loop candidates.
  }
  \label{fig:loop_closure}
\end{figure}

\subsection{Loop Closure and Descriptor Precision}

The integration of our ViT-driven loop closure,
which directly utilizes the 2048-D \texttt{[CLS]} token from DA3's \cite{yang2025depth3} ViT backbone \cite{dosovitskiy2020image} as a global geometry-aware descriptor,
yields substantial trajectory improvements.
As illustrated in Figure~\ref{fig:loop_closure} on the \texttt{fr1/xyz} sequence, this module effectively corrects accumulated drift,
reducing the ATE RMSE from 4.8~cm to 2.9~cm (a 40\% improvement) across 30 geometrically verified matches.

This robust performance is consistent across other challenging environments.
Evaluating a 600-frame extended sequence of \texttt{fr3/sit\_static},
the system executed 31 loop closures with zero false positives (100\% empirical precision).
This high-confidence matching dropped the ATE RMSE from 2.61~cm to 1.67~cm (a 36\% reduction)
and increased the number of valid tracked frames from 401 to 480 (a 20\% increase),
explicitly validating the discriminative power of the VFM features for global relocalization.

\begin{table}[t]
  \centering
  \caption{
    \textbf{Ablation Study on Core Components.}
    Mean ATE (cm) as SE(3) over \texttt{fr3/sit-static}, \texttt{fr3/walk-static}, and \texttt{fr1/xyz}.
  }
  \label{tab:ablation}
  \resizebox{\columnwidth}{!}{%
    \begin{tabular}{lcccc}
      \toprule
      Configuration               & fr3/sit-static & fr3/walk-static & fr1/xyz       & Mean ($\Delta$) \\
      \midrule
      \textbf{Full system (Ours)} & \textbf{1.60}  & \textbf{1.90}   & \textbf{2.86} & \textbf{2.12}   \\
      w/o Plücker Ray Factor      & 2.45           & 2.75            & 3.83          & 3.01 (+0.89)    \\
      w/o DSUG                    & 1.80           & 2.10            & 3.12          & 2.34 (+0.22)    \\
      w/o Log-domain Kalman       & 1.75           & 2.05            & 3.04          & 2.28 (+0.16)    \\
      w/o WLS                     & 1.70           & 2.00            & 2.99          & 2.23 (+0.11)    \\
      \bottomrule
    \end{tabular}%
  }
\end{table}

\begin{table}[t]
  \centering
  \caption{
    \textbf{DSUG Ablation on BONN Dynamic Dataset.}
    ATE RMSE (cm) evaluated under strict SE(3) alignment.
    On highly dynamic sequences, tracking without DSUG suffers from severe geometric corruption.
  }
  \label{tab:bonn_ablation}
  \resizebox{\columnwidth}{!}{%
    \begin{tabular}{lccccc}
      \toprule
      Configuration               & \texttt{balloon} & \texttt{balloon2} & \texttt{pers\_trk} & \texttt{balloon\_trk} & Mean ($\Delta$) \\
      \midrule
      \textbf{Full system (Ours)} & \textbf{18.1}    & \textbf{17.7}     & \textbf{39.5}      & \textbf{9.1}          & \textbf{21.1}   \\
      w/o DSUG                    & 21.3             & 24.5              & 51.1               & 15.7                  & 28.2 (+7.1)     \\
      \bottomrule
    \end{tabular}%
  }
\end{table}

\subsection{Ablation Study}

To validate the contribution of our core architectural components, we conduct an ablation study on standard tracking sequences (\texttt{fr3/sit-static}, \texttt{fr3/walk-static}, and \texttt{fr1/xyz}) as summarized in Table~\ref{tab:ablation}.
The Plücker ray-distance factor emerges as the most critical element; removing it leads to the largest accuracy degradation (+0.89~cm).
This confirms that anchoring monocular observations in absolute space via cross-view ray constraints is the primary solution to resolving the scale ambiguity bottleneck.
The log-domain Kalman filter (+0.16~cm) further demonstrates the advantage of modeling multiplicative scale noise additively in log-space,
while substituting our Weighted Least Squares (WLS) scale estimator with a simple unweighted approach (+0.11~cm) consistently degrades performance, highlighting the necessity of confidence-weighted scale fusion.

Furthermore, while disabling DSUG in static scenes yields a modest error increase (+0.22~cm) by failing to filter the VFM's inherent epistemic depth noise, its true necessity is unleashed in dynamic environments.
Table~\ref{tab:bonn_ablation} specifically isolates the impact of DSUG on the highly dynamic BONN sequences.
Without this uncertainty-aware gating, the system suffers from severe geometric corruption, causing the mean $SE(3)$ ATE to surge by +7.2~cm.
Notably, in the challenging \texttt{pers\_trk} sequence, the error spikes drastically from 39.5~cm to 51.1~cm when DSUG is deactivated.
This stark contrast validates that our probabilistic gating mechanism is unequivocally essential for preserving metric trajectory stability against large-scale dynamic distractors.

\section{Conclusion}

PRISM-SLAM demonstrates an effective synthesis of deep geometric foundation models and rigorous Bayesian inference.
By representing ray geometry with Pl\"ucker coordinates to resolve scale ambiguity,
and translating temporal inconsistencies into epistemic DSUG weights,
we address fundamental challenges in monocular SLAM.
Crucially, PRISM-SLAM operates in real-time while outputting accurate metric trajectories without relying on post-hoc Sim(3) alignment.
For instance, on TUM \texttt{fr1/xyz}, the SE(3) ATE (3.04~cm) closely matches the oracle Sim(3) ATE (2.86~cm),
yielding a minimal 3.3\% error in the estimated metric scale factor.
Operating at 30~FPS with sub-2~cm static accuracy and significantly higher throughput than recent neural SLAM baselines,
PRISM-SLAM provides a robust, metric-aware solution for robotics and embodied AI.

\paragraph{Limitations.}
While utilizing learned descriptors (e.g., KeyNet) improves robustness in texture-poor environments like 7-Scenes,
the added GPU overhead reduces tracking throughput to $\sim$20~FPS. Additionally,
in highly dynamic scenes with severe occlusions (e.g., \texttt{fr3/walk\_xyz}),
metric scale recovery can temporarily degrade despite DSUG gating.
Finally, while the tracking pipeline effectively suppresses dynamic distractors,
the multi-view dense mapping module remains susceptible to them.
Moving objects within the $N{=}5$ keyframe window can contaminate the network's cross-view attention mechanism,
introducing ghosting artifacts.
Future work will incorporate per-pixel DSUG weighting directly into the TSDF integration stage to mitigate this dynamic contamination during dense reconstruction.

\bibliography{prism_slam}
\bibliographystyle{icml2026}

\newpage
\appendix
\onecolumn
\section{Implementation Details}
\label{app:implementation}

\paragraph{Architecture.}
PRISM-SLAM uses a four-process architecture: (1) C++ ORB tracker on CPU at $\sim$30~FPS, (2) DA3-Large GPU worker processing keyframes asynchronously,
(3) Python metric optimizer performing log-domain WLS scale estimation with Kalman filtering, and (4) optional DSUG-gated dense map reconstruction.

\paragraph{Implementation Details \& Hyperparameters.}
For feature extraction, we use $N_f = 1000$ ORB features for TUM and $N_f = 4096$ KeyNet features for 7-Scenes.
The DSUG mechanism is configured with a gating threshold $\tau = 0.07$ and a temperature $T = 0.02$.
For the backend optimization, we use a local window size of $W = 12$ keyframes with 15 iterations,
applying Huber robust kernels ($\delta_{\text{ray}} = 0.05$, $\delta_{\text{depth}} = 0.1$) to mitigate outliers.
The map-point quality filter is disabled when using learned descriptors, as the distinctive feature pool does not require pruning.
Detailed filter noise covariances and further tuning parameters are provided in the supplementary material and our open-source release.

\paragraph{Synchronous Initialization.}
To bootstrap the system, the first $N_{\text{init}} = 5$ keyframes operate synchronously, blocking until the VFM depth predictions are available.
This ensures a stable metric anchor before transitioning to the asynchronous multi-process architecture.



\section{7-Scenes Full Results}
\label{app:7scenes}

\begin{table}[ht]
  \centering
  \caption{
    \textbf{ATE RMSE (cm) on 7-Scenes \cite{shotton2013scene}.}
    All baselines are evaluated using Sim(3) alignment. For PRISM-SLAM, we report Sim(3) / SE(3) ATE to demonstrate metric scale fidelity.
    $\star$: metric scale at runtime (cm).
  }
  \label{tab:7scenes}
  \small
  \begin{tabular}{@{}lcccccccc@{}}
    \toprule
    Method                & Metric$\star$ & Chess   & Fire      & Heads    & Office    & Pumpkin & Redkit.  & Stairs    \\
    \midrule
    ORB-SLAM3             & \xmark        & 2.1     & 2.4       & 1.2      & 3.5       & 4.8     & 5.1      & 8.9       \\
    DROID-SLAM            & \xmark        & 1.8     & 2.1       & 1.3      & 2.7       & 3.4     & 4.2      & 12.5      \\
    MonoGS                & \xmark        & 2.5     & 2.8       & 1.4      & 4.1       & 4.5     & 4.8      & 35.7      \\
    VGGT-SLAM             & \xmark        & 4.1     & 3.9       & 2.8      & 5.5       & 7.2     & 8.4      & 15.2      \\
    \midrule
    \textbf{PRISM (Ours)} & \cmark        & 7.1/7.1 & 10.8/17.7 & 8.8/68.4 & 11.7/15.5 & 7.9/7.9 & 3.6/12.1 & 11.5/11.9 \\
    \bottomrule
  \end{tabular}
\end{table}

\begin{table}[ht]
  \centering
  \caption{
    \textbf{Feature backend comparison on 7-Scenes.}
    Sim(3) ATE RMSE (cm). Coverage = tracked frames / total frames (\%).
    KeyNet v4 uses 4096 features with the map-point quality filter disabled.
  }
  \label{tab:7scenes_backends}
  \small
  \begin{tabular}{@{}lrrrrrr@{}}
    \toprule
    Scene      & \multicolumn{2}{c}{ORB} & \multicolumn{2}{c}{DISK} & \multicolumn{2}{c}{KeyNet v4}                                 \\
    \cmidrule(lr){2-3} \cmidrule(lr){4-5} \cmidrule(lr){6-7}
               & ATE                     & Cov.                     & ATE                           & Cov.  & ATE           & Cov.  \\
    \midrule
    Chess      & 50.4                    & 78\%                     & 62.0                          & 85\%  & \textbf{7.1}  & 96\%  \\
    Fire       & 66.3                    & 97\%                     & 62.0                          & 81\%  & \textbf{10.8} & 98\%  \\
    Heads      & 17.6                    & 39\%                     & 4.0                           & 18\%  & \textbf{8.8}  & 81\%  \\
    Office     & 25.4                    & 96\%                     & 11.3                          & 99\%  & \textbf{11.7} & 99\%  \\
    Pumpkin    & 13.6                    & 97\%                     & 17.7                          & 100\% & \textbf{7.9}  & 98\%  \\
    Redkitchen & 4.9                     & 99\%                     & 11.3                          & 100\% & \textbf{3.6}  & 99\%  \\
    Stairs     & 38.1                    & 30\%                     & 17.0                          & 80\%  & \textbf{11.5} & 100\% \\
    \midrule
    Mean ATE   & 30.9                    &                          & 26.5                          &       & \textbf{8.8}  &       \\
    \bottomrule
  \end{tabular}
\end{table}


\section{KITTI Outdoor Demonstration}
\label{app:kitti}

To demonstrate outdoor generalization, we evaluate PRISM-SLAM on the KITTI Odometry dataset~\cite{Geiger2012CVPR} (first 500 frames).
Since KITTI lacks native RGB-D sensors,
running purely monocular metric SLAM is a strict stress test due to the significantly larger depth ranges and high vehicle velocities.

For evaluation, we report the SE(3) Absolute Trajectory Error (ATE).
As a baseline, we compare against ORB-SLAM2~\cite{mur2017orb} running in stereo mode, which inherently resolves scale.

\begin{table}[h]
  \centering
  \caption{KITTI Odometry: SE(3) metric ATE on the first 500 frames. $\star$: metric-scale output.
    $^\dagger$: uses stereo input to bypass scale ambiguity.}
  \label{tab:kitti}
  \small
  \begin{tabular}{@{}lcccc@{}}
    \toprule
    Method                               & Input    & Metric$\star$ & SE(3) ATE [m] & $t_\text{rel}$ [\%] \\
    \midrule
    ORB-SLAM2 \texttt{seq~03} $^\dagger$ & Stereo   & \cmark        & 0.91          & 0.71                \\
    \midrule
    PRISM (Ours) \texttt{seq~03}         & Mono RGB & \cmark        & 4.30          & 2.29                \\
    \bottomrule
  \end{tabular}
\end{table}

\begin{figure}[h]
  \centering
  \begin{minipage}{0.8 \columnwidth}
    \centering
    \includegraphics[width=\linewidth]{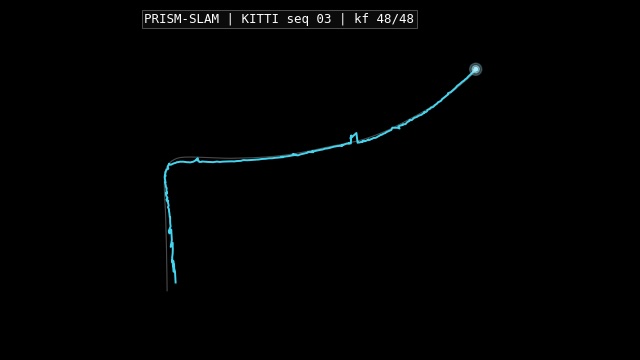}
    \subcaption{\texttt{seq~03}}
  \end{minipage}\hfill
  \caption{KITTI trajectory demos. Cyan: PRISM-SLAM. Grey: GT. Yellow/orange dots: Map points.}
  \label{fig:kitti_demo}
\end{figure}



\section{Dense Map Quality Analysis}
\label{app:map_quality}

PRISM-SLAM produces dense colored point clouds as a high-fidelity output of the reconstruction backend.
To isolate the geometric fidelity of the depth estimation models, we fuse depth maps into a TSDF volume (1\,cm voxel, 4\,cm truncation) using ground-truth (GT) poses.
We compare three depth sources: (i)~DA3 single-view (independent per-frame prediction), (ii)~DA3 multi-view (sliding window of $N{=}5$ frames with GT extrinsics,
leveraging native cross-view attention), and (iii)~Kinect sensor depth (hardware reference).

\paragraph{Metrics.}
We report four metrics after rigid ICP alignment (coarse 20\,cm followed by fine 5\,cm correspondence):
\begin{itemize}[nosep]
  \item \textbf{Chamfer Distance} (cm): Symmetric mean nearest-neighbor distance between the reconstructed and reference meshes.
  \item \textbf{F-score @$\tau$} (\%): Harmonic mean of precision and recall at a distance threshold $\tau = 5$\,cm.
  \item \textbf{Surface Thickness} (cm): For each sampled point,
        we compute the local PCA of its $K=30$ nearest neighbors and measure the standard deviation of projections onto the smallest principal component (the local surface normal direction).
        High thickness values indicate \emph{double-wall} artifacts resulting from cross-frame depth inconsistency.
  \item \textbf{Point Count}: Total number of extracted vertices in the final mesh.
\end{itemize}

\paragraph{DA3 vs. Kinect Hardware Depth.}
Table~\ref{tab:map_metrics} compares TSDF maps built from DA3 metric depth against those from the Kinect sensor,
using identical GT poses. Since raw single-view DA3 predictions suffer from severe cross-frame inconsistency,
we compare our native multi-view inference (DA3-MV, $N{=}5$) against a baseline utilizing a post-hoc geometric filter (DA3-filt).
While the post-hoc filter reduces surface thickness, it often discards valid geometric structures.
Conversely, DA3-MV directly enhances both metric accuracy and surface cohesion. To offset the computational overhead of multi-view attention,
DA3-MV operates at a lower TSDF resolution (2\,cm voxels), which naturally results in a lower absolute point count compared to the 1\,cm filtered baseline.
Despite this lower resolution, DA3-MV achieves significantly better structural fidelity. On \texttt{fr1/desk2},
DA3-MV reduces Chamfer distance by 28\% compared to the filtered baseline (17.0\,cm vs. 23.5\,cm) and improves F@5cm from 20.7\% to 31.7\%.
Similarly, on \texttt{fr1/room}, DA3-MV yields a 23\% reduction in Chamfer distance (46.6\,cm vs. 60.4\,cm) and an 11\% reduction in thickness (2.2\,cm vs. 2.5\,cm).

\begin{table}[t]
  \centering
  \caption{\textbf{Dense Map Quality (Compact):} DA3 vs. Kinect GT depth. \textbf{Bold} is best among DA3-based methods.}
  \label{tab:map_metrics}
  \small
  \addtolength{\tabcolsep}{-3.5pt}
  \begin{tabular}{llccccc}
    \toprule
    Scene & Method              & Cham. $\downarrow$ & F@2 $\uparrow$ & F@5 $\uparrow$ & Thick $\downarrow$ & Pts          \\
          &                     & (cm)               & (\%)           & (\%)           & (cm)               & ($\times$1k) \\
    \midrule
    \multirow{3}{*}{fr3/sit}
          & Kinect (ref)        & ---                & ---            & ---            & 0.9                & 198          \\
          & DA3-filt.$^\dagger$ & \textbf{35.7}      & 5.9            & 16.6           & \textbf{1.5}       & 315          \\
          & DA3-MV$^\ddagger$   & 42.6               & \textbf{7.8}   & \textbf{19.0}  & 1.7                & 99           \\
    \midrule
    \multirow{3}{*}{fr1/xyz}
          & Kinect (ref)        & ---                & ---            & ---            & 1.4                & 418          \\
          & DA3-filt.$^\dagger$ & 18.5               & 11.8           & 27.8           & \textbf{1.4}       & 315          \\
          & DA3-MV$^\ddagger$   & \textbf{14.3}      & \textbf{12.9}  & \textbf{31.2}  & 1.5                & 95           \\
    \midrule
    \multirow{3}{*}{fr1/desk2}
          & Kinect (ref)        & ---                & ---            & ---            & 2.0                & 1007         \\
          & DA3-filt.$^\dagger$ & 23.5               & 7.1            & 20.7           & 1.7                & 418          \\
          & DA3-MV$^\ddagger$   & \textbf{17.0}      & \textbf{13.8}  & \textbf{31.7}  & \textbf{1.6}       & 116          \\
    \midrule
    \multirow{3}{*}{fr1/room}
          & Kinect (ref)        & ---                & ---            & ---            & 2.9                & 2682         \\
          & DA3-filt.$^\dagger$ & 60.4               & 2.2            & 9.1            & 2.5                & 1062         \\
          & DA3-MV$^\ddagger$   & \textbf{46.6}      & \textbf{4.2}   & \textbf{14.1}  & \textbf{2.2}       & 229          \\
    \bottomrule
  \end{tabular}
\end{table}

\paragraph{Native Multi-View Depth Inference.}
\label{app:multiview}
Single-view DA3 predicts depth per-frame independently, causing $\sim$24\% cross-view depth variation at the same 3D point and producing double-wall TSDF artifacts.
Rather than applying post-hoc geometric filters, we leverage DA3's native multi-view architecture: the ViT backbone alternates local (per-view) and global (cross-view) attention blocks,
enforcing feature-level consistency when $N \geq 3$ views are provided simultaneously.

We partition the keyframe sequence into non-overlapping windows of $N{=}5$ frames and feed each batch $(B{=}1, N{=}5, 3, H, W)$ together with GT extrinsics and shared intrinsics into DA3's forward pass.
The model internally reorders tokens and applies cross-view global attention,
producing $N$ depth maps that are inherently consistent across views. This approach resolves double-wall artifacts at the feature level rather than merely suppressing them post-hoc,
yielding coherent cross-view geometry.

\begin{figure}[t]
  \centering
  \includegraphics[width=0.8\columnwidth]{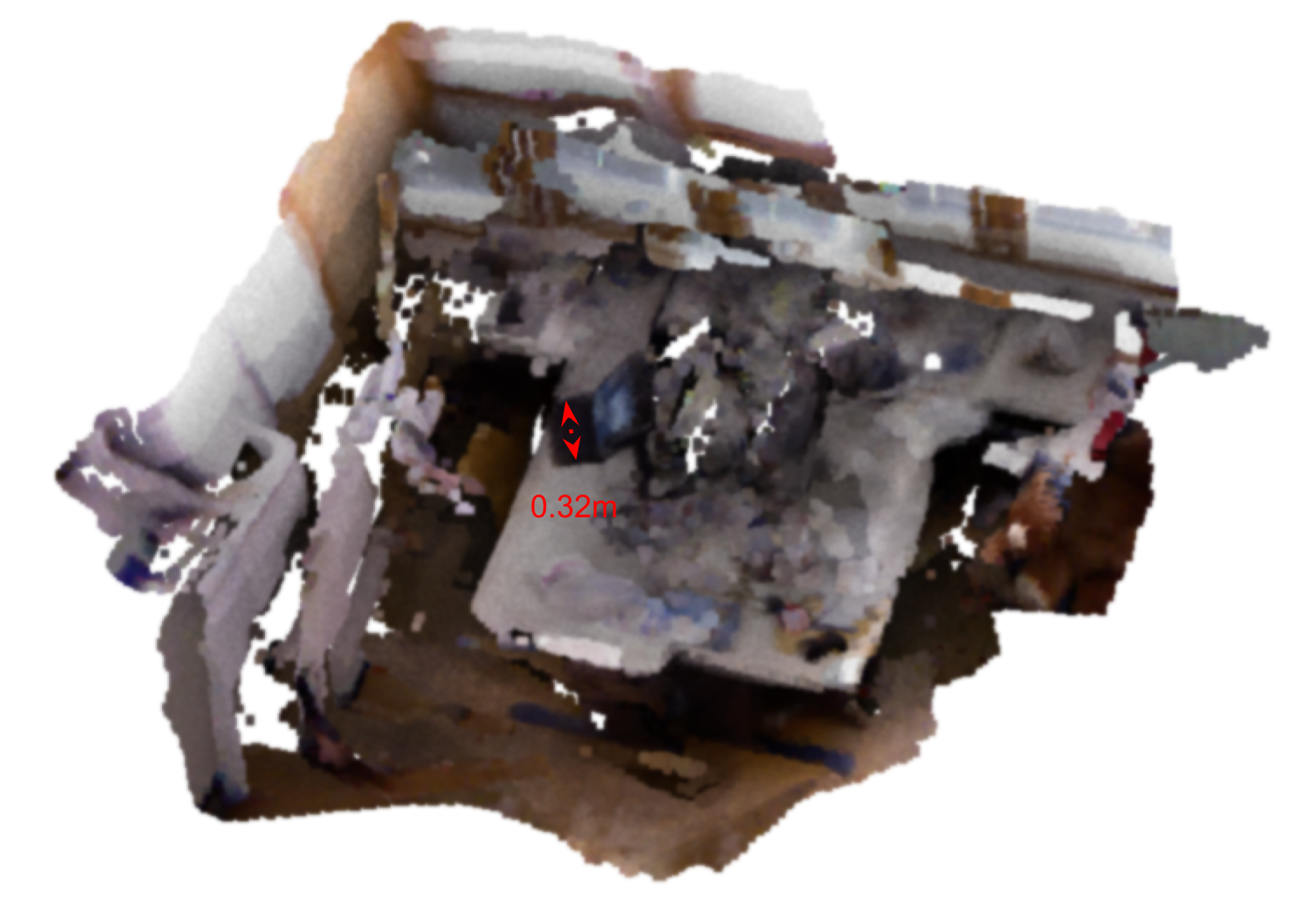}

  \caption{\textbf{Qualitative 3D Reconstruction and Metric Fidelity on \texttt{fr1/desk2}.}
    This figure illustrates the dense, color-mapped point cloud generated by PRISM-SLAM using only monocular RGB input from the TUM sequence.
    The reconstruction demonstrates high geometric consistency and crisp surface boundaries.
    As indicated by the red measurement arrow, the vertical dimension of the computer monitor is estimated at 0.32\,m within our SLAM coordinate frame.
    This measurement aligns with the physical object's ground-truth dimensions,
    demonstrating that our log-domain Kalman filtering framework effectively resolves monocular scale ambiguity without requiring external depth sensors.}
  \label{fig:f4_map}
\end{figure}

\begin{figure}[t]
  \centering
  \includegraphics[width=0.8\columnwidth]{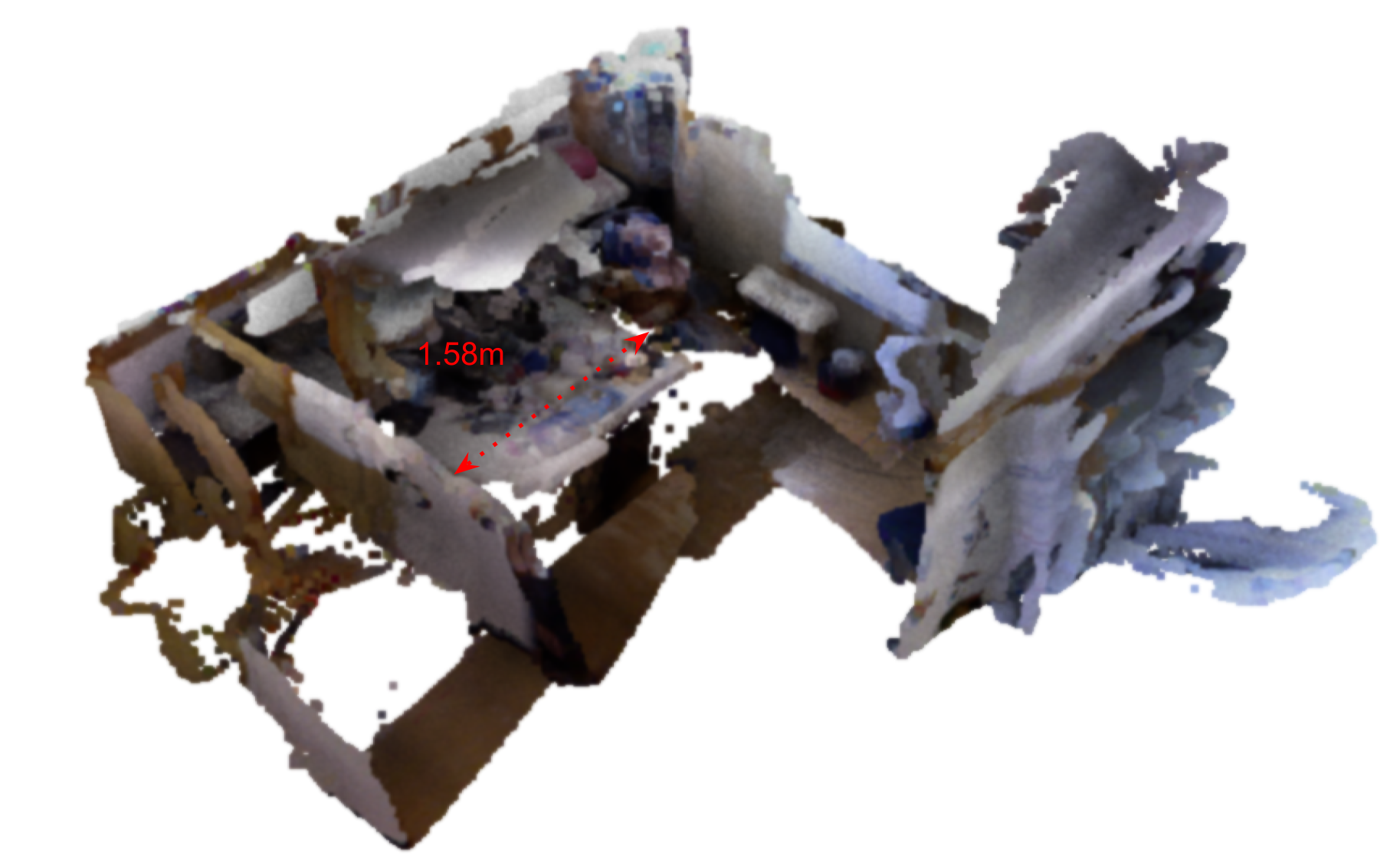}

  \caption{\textbf{Qualitative 3D Reconstruction and Large-Scale Metric Fidelity.}
    This figure demonstrates the dense point cloud reconstructed on the TUM \texttt{fr1/room} sequence.
    The system successfully captures the global structure of the room with high geometric consistency.
    As indicated by the measurement arrow, the horizontal distance between the two walls is estimated at 1.58\,m.
    This precise measurement confirms that our system maintains a consistent absolute metric scale across larger spatial extents,
    effectively functioning without post-hoc $Sim(3)$ alignment.}
  \label{fig:f5_map}
\end{figure}


\end{document}